# REAL-TIME TARGET DETECTION IN MARITIME SCENARIOS BASED ON YOLOV3 MODEL

**Alessandro Betti** [(1)], **Benedetto Michelozzi** [(1)], **Andrea Bracci** [(1)] **and Andrea Masini** [(1)]

[(1)] *Flyby srl, via Aurelio Lampredi 45, Livorno (Italy), Email:* alessandro.betti@flyby.it

**ABSTRACT:**

In this work a novel ships dataset is proposed consisting of more than 56k images of marine vessels collected by means of web-scraping and including 12 ship categories. A YOLOv3 single-stage detector based on Keras API is built on top of this dataset. Current results on four categories (cargo ship, naval ship, oil ship and tug ship) show Average Precision up to 96% for Intersection over Union (IoU) of 0.5 and satisfactory detection performances up to IoU of 0.8. A Data Analytics GUI service based on QT framework and Darknet-53 engine is also implemented in order to simplify the deployment process and analyse massive amount of images even for people without Data Science expertise.

## 1. INTRODUCTION

Real-Time detection of targets in maritime scenarios represents nowadays an essential task to ensure the safety of coast and sea. However, usually ships detection in videos is performed visually, which is both cumbersome and time-consuming. Besides, this aspect is becoming more and more critical as the amount of images acquired by sensors continue to grow, thus demanding the development of automatic detection methods to optimize both response time and object identification accuracy.

Promising results in different domains of target detection have been achieved by means of Convolutional Neural Networks (CNNs) [1] which, unlike Machine Learning methods [2], learn automatically features from input images without requiring human resources and expertise, and enhancing both precision and speed. However, Literature is still poor of public datasets specific for sea ship detection [3, 4], whereas the available ones (ImageNet, COCO, etc.) do not guarantee adequate ships statistics and, therefore, detection performances.

In this paper we present a novel dataset for sea ship detection of 56400 visible images and 12 ship classes acquired both at sea-level and from the sky and obtained as a fusion of different open archives by means of commercial search engines. Images have been annotated by means of high-precision bounding boxes obtained thanks to the open-source LabelImg software [5]. At the best of our knowledge, the proposed dataset is one of the first examples ensuring diversity in terms of background variation, target dimension, viewpoint, lighting conditions and occlusion. Additionally, class unbalancing issues and image size heterogeneity makes our archive a challenging benchmark dataset for achieving satisfactory detection performances.

Moreover we discuss also current results on four categories achieved by means of YOLOv3 network and a GUI developed within the QT framework based on Darknet-53 engine aimed to speed up the machine learning workflow from data preparation to inference and guarantee easy application also for not experienced users.

## 2. DATASET AND FEATURES

### 2.1. Data acquisition

The dataset used for training and inference of our CNN model was built by means of a web scraping procedure of ship images acquired in the visible spectrum. Well-known search engines have been queried: Google, Yandex and Bing. The collected images were captured at either sea-level or from the sky and either onshore or offshore.

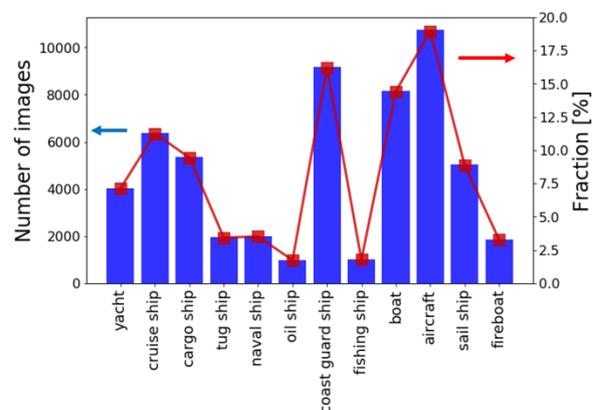

*Figure 1. Dataset statistics. For each category the number of images available (blue bar, left)*

*and the corresponding fraction of the whole dataset (red curve, right) are shown.*

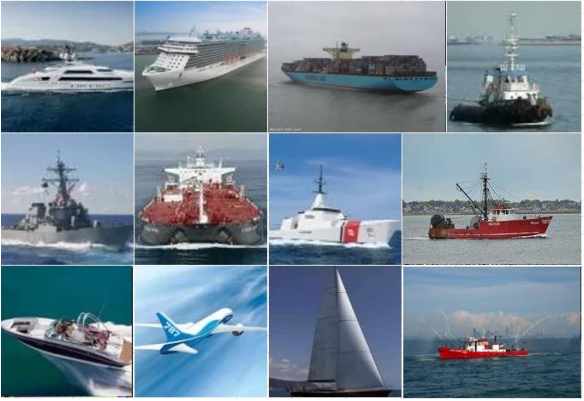

*Figure 2. examples from each category under examination. From top to bottom and from left to right: yacht, cruise ship, cargo ship, tug ship, naval ship, oil ship, coast guard ship, fishing ship, boat, aircraft, sail ship, fireboat.*

A great variety of categories have been included in order to develop a maritime surveillance system suitable for both military and civil application, such as ensuring the safety of coast and the sea or for monitoring marine traffic.
In particular, the archive consists of 12 ship classes: yacht, cruise ship, cargo ship, tug ship, naval ship, oil ship, coast guard ship, fishing ship, boat, aircraft, sail ship and fireboat.
After a pre-processing phase necessary to remove raw images not compliant to the desired categories, the statistics available ranged from almost 1k images (oil ship and fishing ship) up to 11k images (aircraft), thus representing a challenging dataset due to class-unbalancing issues (Figure 1). Globally, the dataset consists of 56400 images. Examples from the different categories are shown in Figure 2.

### 2.2. Dataset variety

Due to the complexity of maritime environment, it is important to collect images in different conditions to guarantee good generalization capability of the CNN model in real-world scenarios.
The following factors were modelled (Figure 3):
i. *Background variation*: in order not to detect the background as part of the target, a variety of onshore and offshore ship images were collected, including vessels in front of coast areas, moored in the harbour and sailing with wakes and waves around.
ii. *Target dimension and visible ship portion*: even if the used YOLOv3 network is able to detect targets at different zoom levels, the availability of ships at different distances simplify the detection process. Additionally, since the CNN must operate also on Real-Time videos, it is mandatory to detect also significant parts of vessels. As a consequence, also images including a part of the ship hull were annotated

iii. *Viewpoint*: in order to apply the CNN model in complex maritime environments which changing scenarios, ship images collected from different points of view and acquired at both sea-level and from aerial vehicles were taken into account
iv. *Lighting conditions*: images captured in different daytimes and weather conditions were included
v. *Occlusion*: usually maritime scenarios include complex situations with more than one ship present and with ship objects partially occluded. Such factor has been carefully examined and included in the dataset.

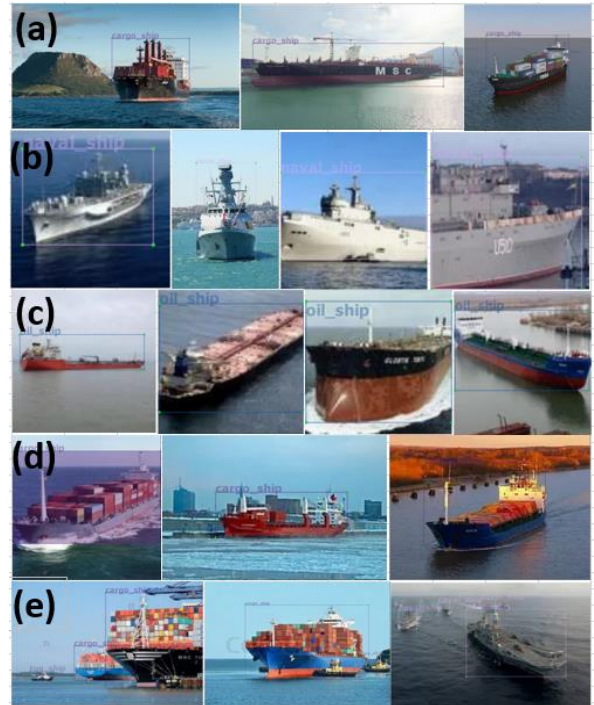

*Figure 3. Environmental factors included in the dataset. From top to bottom and from left to right: (a) background variation (coast, harbour, offshore); (b) target dimension and visible ship portion; (c) viewpoint; (d) lighting conditions (sunny day, cloudy day and at the sunset); (e) occlusion.*

### 2.3. Dataset annotation

Image annotation is a demanding process typically performed by humans and necessary to train the CNN model to classify and localize the correct targets in the images. In this work two different annotation methods have been used:
i. First, the images returned by web scraping have been grouped according to the category of the main foreground object in the frame. Then for each category the images have been divided in batches of almost 500 images each and assigned to the company staff for preliminary manual annotation.
ii. Once sufficient training statistics was available, semi-automatic annotation mode has been adopted on new images by inferring ship location and label by means of a CNN partially finetuned on the category under investigation. Human supervision was however still necessary to verify the quality of annotated images and fix ground truth boxes or class labels if incorrect.

In this work the LabelImg annotation tool [5] has been used by drawing a tight bounding box around the ship target and labelling the box according to the corresponding ship category. The generated txt file follows the YOLO format shown in Table 1 with a new line for each target included in the image: class_ID is the ship category label, $t_x$ ($t_y$) is the x (y) coordinate of the bounding box centroid, whereas $w_{gt}$ ($h_{gt}$) is the ground-truth box width (height).

| class_ID | $t_x$ | $t_y$ | $w_{gt}$ | $h_{gt}$ |
|---|---|---|---|---|

*Table 1. The entries saved in the annotation file with YOLO format.*

### 2.4. Explorative Data Analysis of the annotated dataset

Due to the large amount of images to label, the process is still ongoing and batches corresponding to four categories have been annotated: cargo ship, naval ship, oil ship and tug ship. As can be seen in Figure 4, at the moment a number of instances larger than 2k is available for classes naval ship and cargo ship, whereas oil ship and tug ship classes consists of around 1k objects each. Globally, 5792 annotated images are available including more than 7000 ground truth boxes. Furthermore, since many images include more than one ship sailing, i.e. not only the foreground ship object, the actual annotated dataset is not limited to these 4 categories but includes also background instances of other classes, as it is shown in Figure 4. In particular, almost 350 aircrafts are available due to the presence of aircraft carrier images, whereas for the other classes statistics is insufficient for training a Deep Learning architecture.

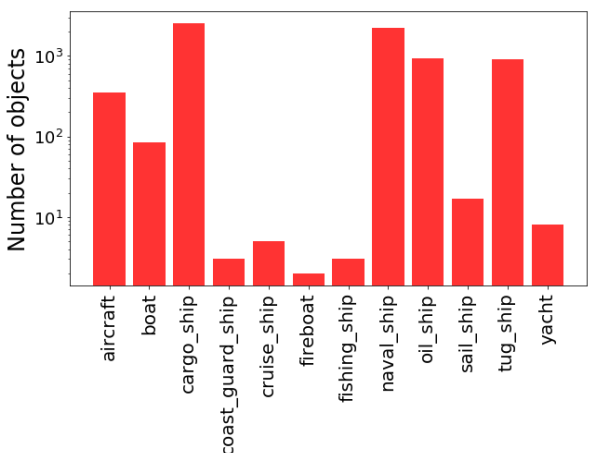

*Figure 4. Number of ship objects available for the different categories in the annotated dataset.*

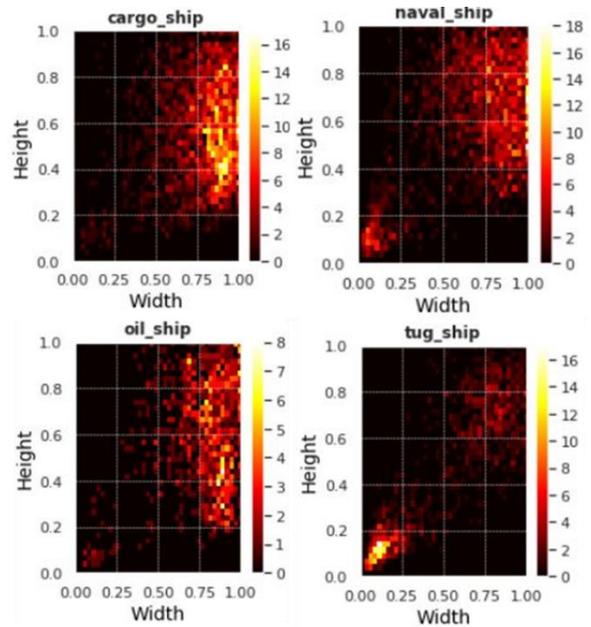

*Figure 5. 2D density plot of ground truth objects in the plane ($w_{gt}$, $h_{gt}$). From left to right and from top to bottom the categories cargo ship, naval ship, oil ship and tug ship are shown.*

Additionally, almost 400 tug ships out of the 900 available are background objects in cargo and oil ship images.

*Figure 5* shows the 2D density plot of ground truth objects in the plane defined by their dimension ($w_{gt}$, $h_{gt}$). As expected, cargo and oil ships tend to extend more in width than in height, with limited statistics at small scale. On the contrary, naval ships, and especially tug ships, exhibit a very significant amount of tiny objects with dimension smaller than 10-20% of the whole image on both sides.

In Figure 6 the 2D density plot of ground truth objects in the plane defined by the Aspect Ratio (AR), i.e. $AR = w_{gt} / h_{gt}$, and normalized target area, i.e. $A_{gt} = w_{gt} \cdot h_{gt}/(W \cdot H)$ , is instead shown. Here W (H) is the image width (height). AR as large as 6 or 7 can be achieved for cargo and oil ships, with A ranging mainly from 25% up to 100% of the full image area. Unlike oil tanker, naval and tug ships exhibit an AR usually smaller than 3 and with a remarkable fraction of targets having $A_{gt} \leq 10\%$.

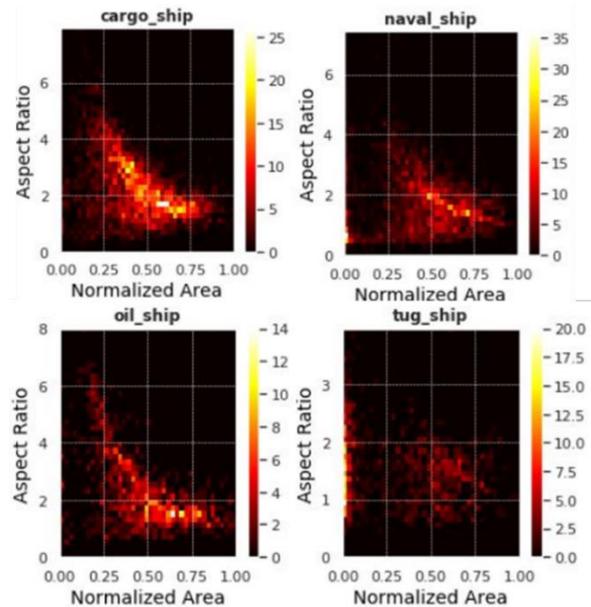

*Figure 6. 2D density plot of ground truth objects in the plane (AR, $A_{gt}$). From left to right and from top to bottom the categories cargo ship, naval ship, oil ship and tug ship are shown.*

## 3. METHODS

### 3.1. The Keras Model based on YOLOv3

A YOLOv3 single-stage multi-scale detector has been re-trained to verify the applicability of the proposed dataset for Real-Time detection in maritime environments. In particular, a Keras model on top of Tensorflow backend has been designed and tested on a Microsoft Azure Virtual Machine specifically customized for Deep Learning applications. As known [6], YOLOv3 is composed by a backbone and a head subnet. The backbone is represented by the Darknet-53 network and contains 1x1 and 3x3 convolutional filters and is responsible for computing the feature maps over the entire input image. The head subnet is built on top of the backbone and is based on a Feature Pyramid Network [7] which alleviate the problem of small targets detection by implementing detection at three different scales and performing both classification and bounding box regression. In particular, the input image is automatically resized to a default size (416x416 in our case) and divided in a grid, with each grid cell producing in output an array whose shape is *BX(5+C)* where B is the number of bounding boxes a cell can predict, 5 is for the number of bounding box attributes and the object confidence, and C is the number of classes. Non-Maximal Suppression is finally used to only keep the predicted bounding boxes having highest confidence.

To optimally exploit the available yet limited dataset, it has been divided in training and test sets according roughly to the proportions 75% and 25%, respectively, for each of the 4 main ship categories (Table 2). Three anchor boxes have been considered for the three different scales and priors of anchors are computed by means of unsupervised k-means clustering.

In Literature Data Augmentation techniques are poorly applied [8] due to unpredictability of distortions present in real scenarios. In this work, due to the limited ground truth data, we implemented it on training images in the forms of scaling, with the constraint of unchanged AR, and horizontal flipping. These methods enrich training set while still yielding realistic images. Image distortion was instead not used since providing unrealistic colours.

We have assumed a typical two-stage pipeline for training of YOLOv3 architecture:
i. *Transfer Learning*: Darknet-53 backbone is pre-trained on the ImageNet dataset. Due to the presence of the ship class also in ImageNet, highest level features extracted from such dataset are used by unfreezing only the last three convolutional layers of Darknet-53, which are instead trained on the new dataset
ii. *Finetuning*: the entire architecture is finetuned starting from weights achieved after 50 epochs of transfer learning

Training is executed by minimizing loss function according to the stochastic Adam optimizer and using an initial learning rate of $10^{-4}$. Early stopping method is finally implemented to avoid overfitting on training set by randomly putting aside 10% of training instances for validation purpose. Once training is early stopped, the detector is able to infer on test set and on videos almost 8-10 frames per second (fps).

Performances are finally evaluated for each ship class in terms of Average Precision (AP), defined as the area under the Precision-Recall curves, for different Intersection over Union (IoU) between the ground-truth box and the predicted one. Furthermore, since for each ship type the dataset is unbalanced towards the negative class (i.e. the class including all the other ship types), we monitor singularly the quantities False Negative rate, i.e. FNR= FN/P= 1-Recall, and FP/P, where FN (FP) is the number of False Negatives (False Positives) and P= TP+FN is the overall amount of Positive samples (TP: True Positive). Overall performance is finally evaluated by means of the mean Average Precision (mAP), defined as $mAP = 1/n \cdot \sum_{i=1}^{n} AP_i$ and the sum runs over the n= 4 ship types.

| Ship class | Training set | Test set |
|---|---|---|
| Cargo ship | 1881 | 633 |
| Naval ship | 1691 | 515 |
| Oil ship | 779 | 149 |
| Tug ship | 707 | 207 |

*Table 2. Number of ground-truth objects in training and test sets for each ship type.*

### 3.2. The training GUI based on Darknet-53 engine and the SDK library

In order to simplify the workflow process from data exploration up to model creation a GUI based on Darknet-53 engine has been also designed within the QT framework. In particular the user may easily import the input dataset with preview capabilities, compute anchor boxes on training set, make a training session visualizing the loss function progress and execute inference returning a detection performances report. Additionally, with the aim to enhance further the detection speed and allow the design of customized applications, an SDK library has been also released which couples the detection capabilities of Darknet-53 with a tracking algorithm based on the Lucas-Kanade optical flow method. In this way the software may track the target in intermediate frames which cannot be processed by the detector due to limited fps.

## 4. RESULTS AND DISCUSSION

### 4.1. The Keras Model based on YOLOv3

In Figure 7a and b (Figure 7c and d) the Precision-Recall curves and the amount of TP, FP and FN are shown, respectively, for each class @IoU= 0.5 (0.8). Figure 8 shows instead AP for the four ship types, as well as mAP (black curve), as a function of IoU. As can be seen in Figure 8, mAP is almost 86% @IoU= 0.5 and remains above 80% up to IoU= 0.7.

Concerning specific ship types, performances for naval ships are far better than for the other classes with AP as large as 96% @IoU= 0.5 (Figure 7a) and decreasing slowly even for larger IoU (AP > 87% @IoU= 0.8). Both Precision and Recall are very satisfactory @IoU= 0.5 with false alarm ratios FP/P and FN/P as small as 3-4% (Figure 7b). This can be explained with significant $A_{gt}$ of naval ships and their distinctive features such as the gray colour.

Remarkable performances are also obtained for cargo and tug ships, with AP almost equal to 92% and 85%, respectively, @IoU= 0.5 and with a degradation rate similar up to IoU= 0.8, where AP is around 60-65% for both.

Additionally @IoU= 0.5 misdetections tend to be

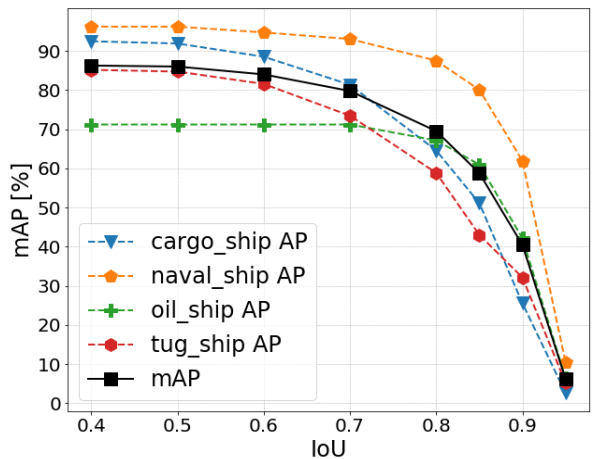

Figure 8. AP as a function of IoU for cargo (blue), naval (orange), oil (green) and tug (red) ships. mAP (black curve) is also shown.

more frequent than missed detections (FP/P= 10% for cargo and 17% for tug), with a modest FN/P around 6% for cargo ship (Figure 7b). @IoU= 0.8 performances are instead degraded yet similar for both class with values ranging from 25% (FN/P, cargo ship) up to 39% (FP/P, tug ship), indicating some lacks in localization capability (Figure 7d).

Good results @IoU= 0.5 for cargo ships may be explained with their distinctive features, such as large AR and their use for containers transportation. On the contrary, tug ships have often a limited $A_{gt}$, especially when around cargo ships, and AR. However their characteristics are significantly different from the other ship types considered and this may explain the satisfactory results even for small targets.

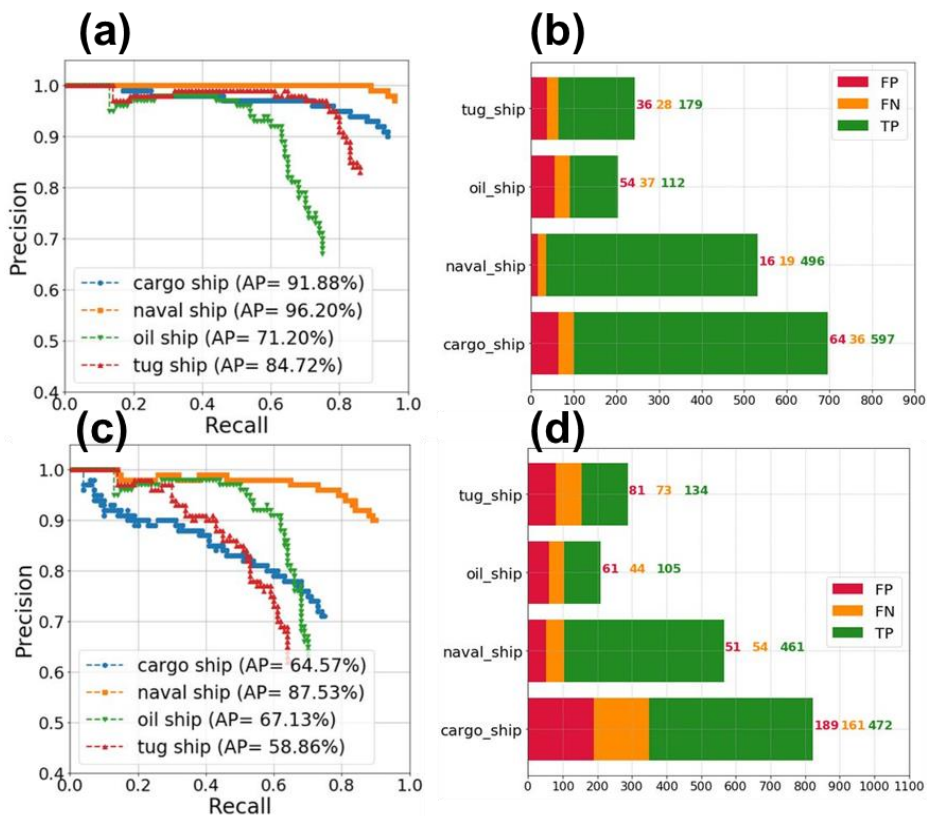

Figure 7. (a), (c) Precision-Recall curves and (b), (d) number of TP, FP and FN for the four ship categories. (a), (b) are evaluated @IoU= 0.5, whereas(c), (d) @IoU= 0.8.

Detection for oil ships is instead worse than other categories even for low IoU (AP= 71% @IoU= 0.5), indicating modest classification performances rather than localization ones. This is confirmed by the slow degradation rate at high IoU if compared to the other ship types (Figure 8). In particular the model is biased towards the prediction of FP (FP/P= 36%) than FN (FN/P= 25%). The main reason for this is their similarity with cargo ships in terms of AR and $A_{gt}$, as seen in Figure 6, which are often confused as oil ships. Furthermore, they are almost indistinguishable if cargo ship does not transport containers but goods inside the hull.

Figure 9 shows some examples of detection results for the different ship types achieved in different situations: from left to right small ship target, complex scenario with some ships, limited visible ship portion and target occlusion. As already highlighted, performances for naval ships are remarkable even for small targets (Figure 9a) or targets very close (Figure 9d).

Results are very good also for cargo ship, at least until they transport containers. In this case, it is sufficient the bow part to detect the ship (Figure 9g).

YOLOv3 demonstrates also satisfactory performances in localizing small targets such as tug ships (Figure 9i and j), and also in conditions of partial ship occlusions and challenging dark background (Figure 9l). On the other hand, some missed detections still occur (Figure 9b).

As already observed, at the moment results for oil ships are perfectible. While the model is able to detect targets also far from the camera (Figure 9m) or in images with limited ship portion available (Figure 9o), a consistent amount of misdetections or missed detections is triggered due to similarity with cargo ships (Figure 9n). FN are also generated in correspondence of very close targets (Figure 9p).

Impressively, even if the aircrafts statistics is limited, YOLOv3 can already detect a significant number of objects (Figure 9d). This may be due to the use of pre-trained weights on ImageNet dataset, which already includes the aircraft class.

### 4.2. The training GUI based on Darknet-53 engine and the SDK library

Our effort was also oriented towards the development of a GUI simplifying the deployments process of a Deep Learning model from image annotation and dataset exploration up to model inference. In particular the following functionalities have been implemented:

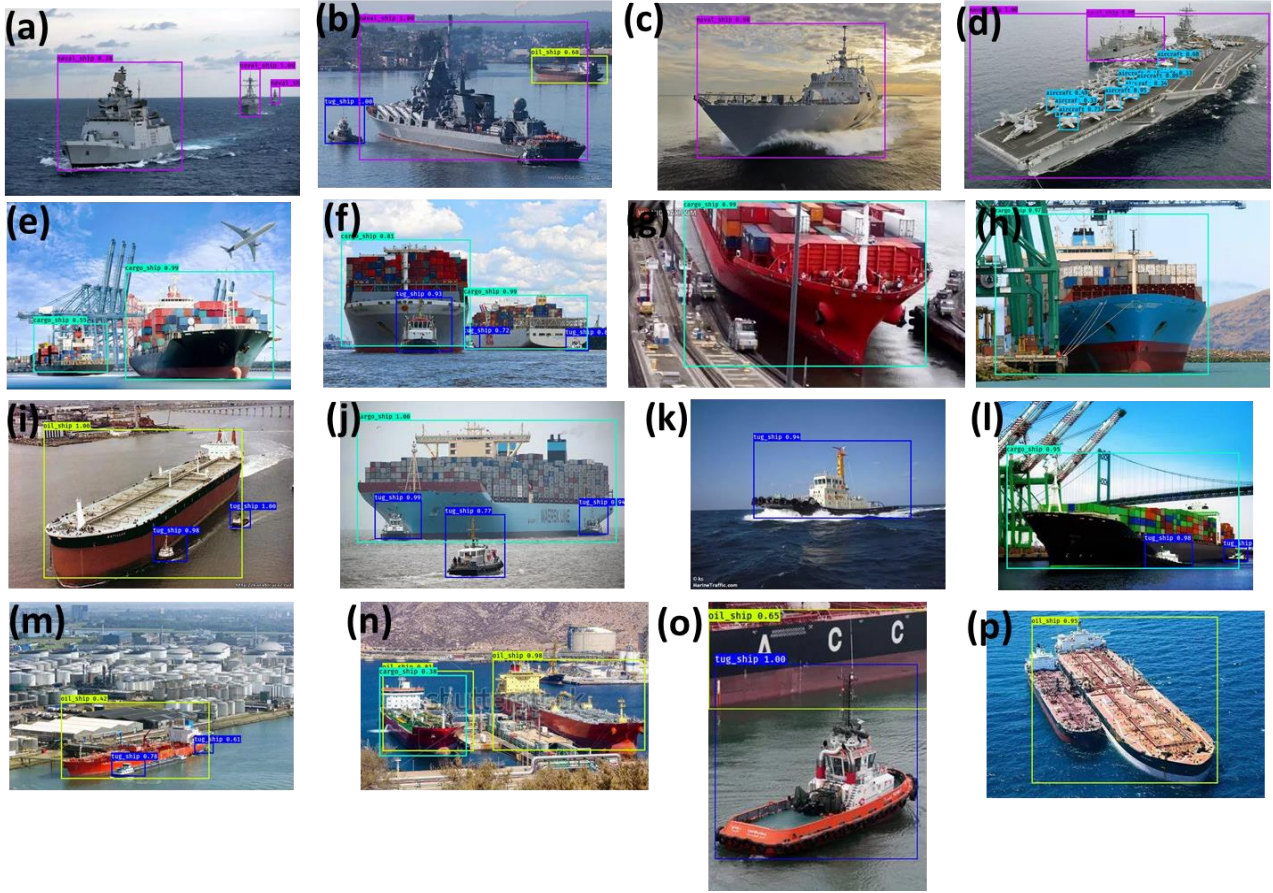

*Figure 9. Ship detection results. From left to right the following situations are considered: small targets, complex scenario with different targets, limited visible ship portion and target occlusion. From top to bottom the following ship classes are shown: naval ship, cargo ship, tug ship and oil ship.*

*Figure 10. Training mode.*

*Figure 11. Inference mode (one-shot).*

*Figure 12. The SDK library.*

i. image annotation (based on LabelImg)
   ii. dataset exploration
   iii. anchor box computation
   iv. training (finetuning)
   v. inference either of an image folder or of a single image (one-shot)

In Figure 10 the training mode is presented. Training may be executed in finetuning mode by selecting either pre-trained ImageNet weights or custom weights. User may inspect the imported annotated dataset and the simulation parameters on the top left and on the bottom left panels, respectively. Furthermore, he can monitor the training status either in the GUI console (bottom right panel) or visually in a line plot showing the loss function progress (top right panel).

Figure 11 shows instead the inference mode starting from the trained CNN. In particular the GUI allow to compare the ground truth object (top left panel) and the detection results (top right panel) easily, simplifying performance validation. Simulation progress is instead shown on the bottom right console, as usual. Finally in Figure 12 the use of the SDK library to track targets on a video is shown, returning also the ship class.

## 5. CONCLUSIONS

In conclusion, in this work (i) we have proposed a novel extended ship dataset suitable for benchmarking activities, modelling a list of 12 ship classes with the aim of supporting the detection of every target may be encountered in maritime scenarios; (ii) a Real-Time YOLOv3 detector has been designed on top of this dataset to discuss the main difficulties of ship detection and demonstrate that YOLOv3 is promising in capturing discriminative features of ship targets; (iii) a user-friendly GUI based on Darknet-53 engine has been also designed to speed up the deployment process even for not experienced users; (iv) an inference module based on Darknet-53 and Lucas-Kanade optical flow algorithm is finally released as an SDK library in order to allow the inclusion into complex software architecture.

Next steps include the extension of annotation to the full dataset and inclusion of images from real-world videos to manage specific conditions infrequent in web-scraping data. Furthermore, use of rotated bounding boxes may help in maximizing localization skills, especially for images collected by aerial vehicles [7]. Finally comparison of the proposed dataset with other open archives [4] and test on Real-Time videos is necessary to verify the generalization capability of the proposed model in different contexts and in operative scenarios.

## 6. ACKNOWLEDGEMENTS

This article is a follow-on of results presented at the EDA workshop on "Artificial Intelligence for OPTRONICS systems" held at Brussels (2018). Results achieved in this work are exploited in the project SPIDVE for design of new methodologies for EO Sensors Performance Improvement in Degraded Visual Environment.
The authors acknowledge the company staff for support in the image annotation process.

## 7. REFERENCES

1. X. Yang, H. Sun, K, Fu, J. Yang et al (2018). Automatic Ship Detection of Remote Sensing Images from Google Earth in Complex Scenes Based on Multi-Scale Rotation Dense Feature Pyramid Networks. *Remote Sensing*, 10, 132.
2. O. Atalar, B. Bartan (2018). Ship Classification Using an Image Dataset. 21st International Conference on Information Fusion (FUSION), At Cambridge, UK
3. Y.-L. Chang, A. Anagaw, L. Chang, Y. C. Wang et al (2019). Ship Detection Based on YOLOv2 for SAR Imagery. *MDPI Remote Sensing*, 11, 786
4. Z. Shao, W. Wu, Z. Wang, W. Du et al (2018). SeaShips: A Large- Scale Precisely Annotated Dataset for Ship Detection. *IEEE Transaction on Multimedia*, 20, 2593.
5. https://github.com/tzutalin/labelImg
6. J. Redmon, A. Farhadi (2018). YOLOv3: An Incremental Improvement. Tech Report.
7. T.Y. Lin, P. Dollar, R. Girshick, K. He et al (2016). Feature pyramid networks for object detection. IEEE Conference on Computer Vision and Pattern Recognition, Honolulu, HI, 936-944
8. C. M. Ward, J. Harguess, C. Hilton (2019). Ship classification from overhead imagery using synthetic data and domain adaption. OCEANS 2018 MTS/IEEE Charleston